\documentclass{article}

\usepackage{PRIMEarxiv}

\usepackage[utf8]{inputenc} 
\usepackage[T1]{fontenc}    
\usepackage{hyperref}       
\usepackage{url}            
\usepackage{booktabs}       
\usepackage{amsfonts}       
\usepackage{nicefrac}       
\usepackage{microtype}      
\usepackage{lipsum}
\usepackage{fancyhdr}       
\usepackage{graphicx}       
\graphicspath{{media/}}     
\usepackage{algorithm}
\usepackage{algpseudocode}
\usepackage{amsmath}
\usepackage{amssymb}
\usepackage{bm,times}
\usepackage{subcaption}
\usepackage{multirow}
\title{Mitigating Viewer Impact from Disturbing Imagery using AI Filters: A User-Study}

\author{
\name{Ioannis Sarridis\textsuperscript{a}\thanks{CONTACT Ioannis Sarridis. Email: gsarridis@iti.gr}, Jochen Spangenberg\textsuperscript{b}, Olga Papadopoulou\textsuperscript{a} and Symeon Papadopoulos\textsuperscript{a}}
\affil{\textsuperscript{a}Centre for Research and Technology Hellas, 6th km Charilaou-Thermi Rd, Thessaloniki 57001, Greece; \textsuperscript{b}Deutsche Welle, Voltastr. 6,
13355 Berlin, Germany }
}

\author{%
  Ioannis Sarridis \\
  Centre for Research and Technology Hellas\\
  Thessaloniki, Greece\\
  \texttt{gsarridis@iti.gr} \\
  \And
   Jochen Spangenberg \\
   Deutsche Welle\\
   Berlin, Germany  \\
   \texttt{jochen.spangenberg@dw.com} \\
   \And
   ~~~~~Olga Papadopoulou \\
   ~~~~~Centre for Research and Technology Hellas \\
   ~~~~~Thessaloniki, Greece \\
   ~~~~~\texttt{olgapapa@iti.gr} \\
   \And
   Symeon Papadopoulos \\
   Centre for Research and Technology Hellas \\
   Thessaloniki, Greece \\
   \texttt{papadop@iti.gr} \\
}

\begin{document}
\maketitle

\begin{abstract}
Exposure to disturbing imagery can significantly impact individuals, especially professionals who encounter such content as part of their work. This paper presents a user study, involving 107 participants, predominantly journalists and human rights investigators, that explores the capability of Artificial Intelligence (AI)-based image filters to potentially mitigate the emotional impact of viewing such disturbing content. We tested five different filter styles, both traditional (Blurring and Partial Blurring) and AI-based (Drawing, Colored Drawing, and Painting), and measured their effectiveness in terms of conveying image information while reducing emotional distress. Our findings suggest that the AI-based Drawing style filter demonstrates the best performance, offering a promising solution for reducing negative feelings (-30.38\%) while preserving the interpretability of the image (97.19\%). Despite the requirement for many professionals to eventually inspect the original images, participants suggested potential strategies for integrating AI filters into their workflow, such as using AI filters as an initial, preparatory step before viewing the original image. Overall, this paper contributes to the development of a more ethically considerate and effective visual environment for professionals routinely engaging with potentially disturbing imagery.

\end{abstract}

%

\section{Introduction}
In the era of digital communication there is an exponential increase in media content, with potentially disturbing and traumatizing images becoming increasingly prevalent \cite{dubberley2015making, zeng2018danger,reid2014journalists}. This issue holds particular significance for professions such as journalism and human rights investigation, where interactions with distressing visual content are occupational inevitabilities \cite{dubberley2015making}.
Such graphic visuals frequently encapsulate scenes of violence, harm, and suffering, provoking emotions of worry, concern, or anxiety that can lead to secondary or vicarious trauma \cite{zampoglou2016web, sarridis2022leveraging,shah2023}.
For instance, professionals may be required to inspect footage from conflict zones such as the war in Ukraine \cite{spangenberg2022}, scenes from natural disasters, or horrific accidents.
Thus, it is crucial to develop and employ solutions that can effectively mitigate the viewer's impact from such disturbing imagery.

Conventional solutions to this issue have predominantly focused on the application of traditional image filters, such as blurring \cite{das2020fast,giancarlo2022}. However, these traditional filters come with significant drawbacks. If applied too heavily, blurring can render an image virtually unrecognizable, stripping away essential details and making the content impossible to interpret \cite{karunakaran2019testing}. If not enough distortion is applied, however, disturbing elements are not sufficiently masked, thus failing to mitigate the negative impact on the viewer. This creates a challenging trade-off between the preservation of information and the protection of the viewer.

The rapid advancements of Artificial Intelligence (AI) in recent years have enabled its integration into numerous fields with a wide range of applications \cite{taigman2014deepface,creswell2018generative,bobadilla2013recommender,tan2020efficientdet}. 
Among the areas where AI systems have exhibited notable effectiveness is the neural image style transfer \cite{gatys2016image,jing2019neural,deng2022stytr2,luo2022progressive}, i.e., the process of altering digital images to adopt the appearance or visual style of another image.

\begin{figure}[t]
\begin{tabular}{ccc}
\centering
  \includegraphics[width=0.3\linewidth]{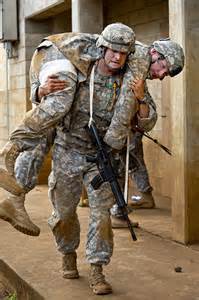} &   
  \includegraphics[width=0.3\linewidth]{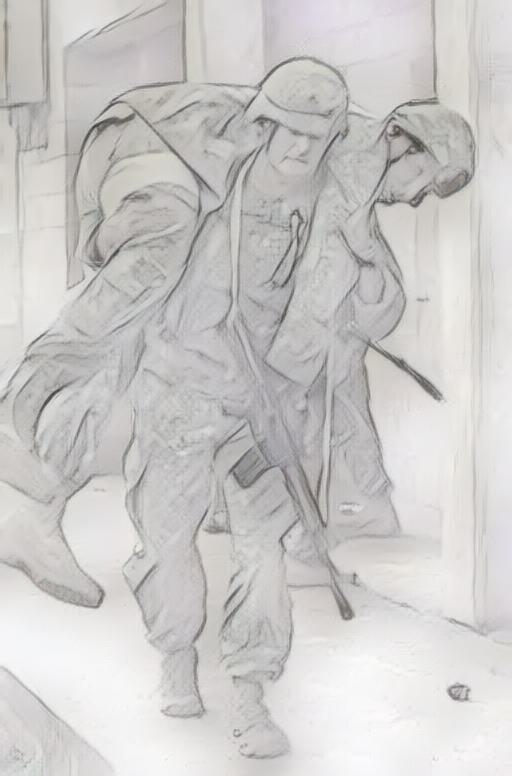} &
  \includegraphics[width=0.3\linewidth]{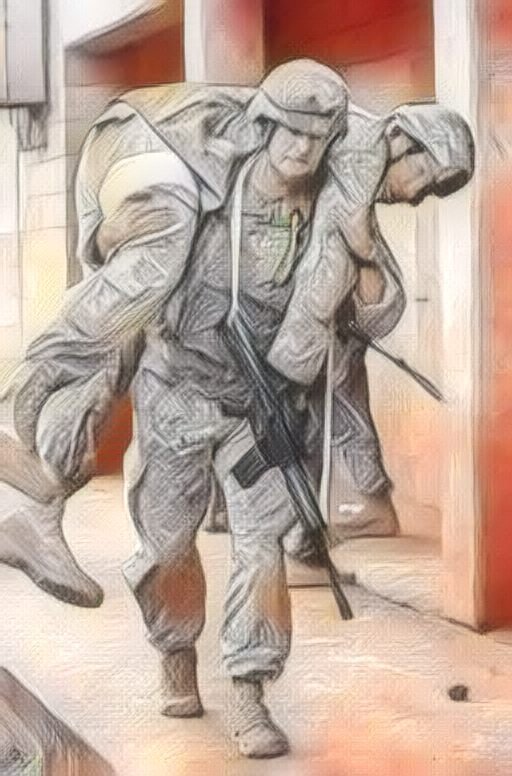} \\    
  (a) Original & (b) Drawing & (c) Colored Drawing  \\
  \includegraphics[width=0.3\linewidth]{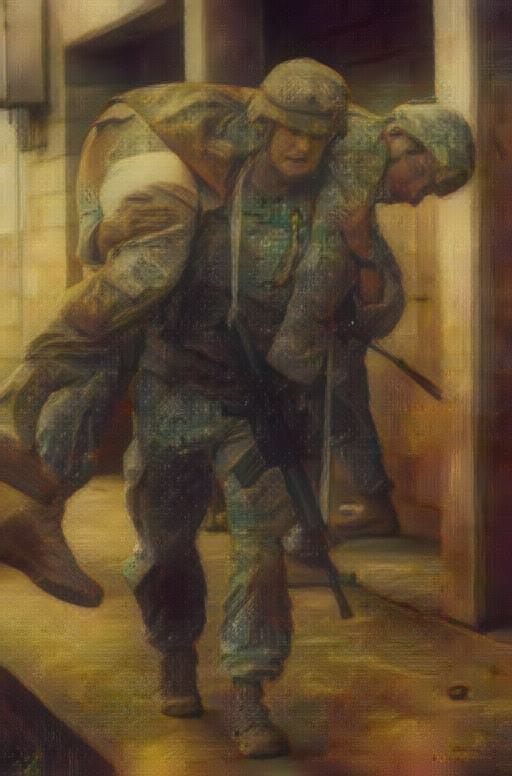} &
  \includegraphics[width=0.3\linewidth]{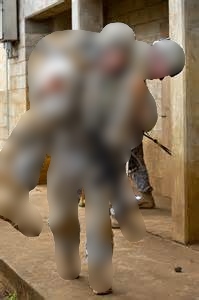} & 
  \includegraphics[width=0.3\linewidth]{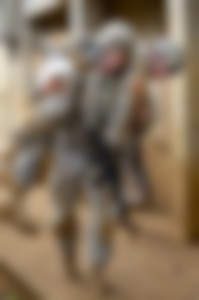}\\
(d) Painting & (e) Partial Blurring & (d) Blurring \\
\end{tabular}
\caption{Examples of AI-based and conventional filters.}
\label{fig:example}
\end{figure}

In this paper, we investigate the potential of AI style-transfer filters to mitigate the distressing impact of graphic imagery, thereby addressing the inherent limitations of conventional blurring techniques. To this end, we have adopted three distinct styles/filters, i.e., Drawing, Colored Drawing, and Painting (see Figure~\ref{fig:example}), and conducted a user study to compare their effectiveness with that of traditional filters. 
It is important to stress that this study focuses on images containing explicit scenes of violence, injury, etc. It does not aim to detect or address every potentially traumatizing or distressing content due to the diverse range of triggers that different individuals may have. For instance, an image featuring a sorrowful child could potentially evoke distress, yet such images cannot easily be identified as potential distress triggers.

The 107 participants of this study (details about study set-up in Section~\ref{sec:method}) are individuals from professional fields that often entail regular engagement with distressing digital content (e.g., journalists, investigators, etc.). The conducted evaluation is primarily based on two key axes - the intensity of the negative emotional responses triggered while viewing the filtered images and the degree of information retained within these images. The latter is of high importance in the relevant professional contexts where detail identification 
is essential.

The findings of this user study confirmed the potential of AI filters to protect the mental well-being of such professionals. In particular, it was observed that compared to the conventional Blurring filters, the Drawing filter was more effective in reducing the negative emotional impact of viewing distressing images, as evidenced by the lower mean ratings used to measure negative feelings (i.e., -34.14\%). In addition, Drawing maintained a significant amount of image detail (97.19\%) necessary for various professional purposes, which is not the case for the Blurring filter (6.54\%). It is worth noting that the absence of color and the regional consistency of the Drawing filter were the two major advantages compared to the other filters.
Furthermore, feedback from participants indicated a broad acknowledgment of the potential utility of AI filters in their professional contexts. 
They highlighted specific stages in their workflow where such filters could be beneficially incorporated, proposed additional enhancements that could facilitate this integration, and noted potential limitations.
The main contributions of this paper are the following:
\begin{itemize}
    \item Exploring the application of AI style transfer filters as a valuable tool for mitigating the emotional impact caused by disturbing digital content, with a focus on professions such as journalism and human rights investigation, where exposure to distressing imagery is a routine occurrence.
    \item  A comprehensive user study comparing the effectiveness of AI-based filters against traditional blurring techniques. The results indicate the promising performance of the AI style transfer filters.
    \item Presenting user feedback, detailing potential workflow integration points, potential improvements, and limitations. 
\end{itemize}
The remainder of this paper is organized as follows: Section~\ref{sec:related} provides an overview of related work, Section~\ref{sec:method} presents the methodology followed for the user study conducted, and the results of the performed analysis are detailed in Section~\ref{sec:results}. Finally, we conclude with Section~\ref{sec:conc}, summarizing our findings, outlining the study's limitations, and suggesting directions for future research.

\section{Related Works}
\label{sec:related}
\textbf{Professions Associated with Exposure to Disturbing Digital Imagery}.
The exposure to disturbing user-generated content (UGC) has been recognized as a significant issue across multiple professions, including journalism, human rights investigations, content moderation, and criminal justice, among others. Zeng et al.
\cite{zeng2018danger} delve into the ethical responsibilities of news organizations towards journalists processing UGC, emphasizing the risk of secondary trauma and Post-Traumatic Stress Disorder (PTSD) symptoms. Similarly, the authors of a study conducted for Eyewitness Media Hub\footnote{http://eyewitnessmediahub.com/} highlight that journalists engaged in the verification and editing of traumatic UGC can suffer `secondary trauma' and symptoms associated with PTSD \cite{dubberley2015making}. Feinsteing et al. \cite{feinstein2014witnessing} and Reid \cite{reid2014journalists} also align with this viewpoint, suggesting that the frequency or duration of exposure to graphic imagery escalates the likelihood of vicarious trauma. These studies recommend protective measures such as staff rotation, peer support, and preemptive hiring warnings.
Hill et al. \cite{hill2020preparing} and Baker et al. \cite{baker2020safer} further emphasize the emotional impact of reporting on traumatic events and reviewing graphic war crime imagery for journalists and human rights investigators, respectively. Both studies highlight the risk of secondary trauma and the need for strategies to mitigate this risk.
Pearson et al. \cite{pearson2023online} provide insight into the harms experienced by online extremism and terrorism researchers due to their exposure to distressing content.
Furthermore, psychological traumas on content moderators are highlighted in several studies \cite{arsht2018human, steiger2021psychological}.
Finally, in-depth interviews with human content moderators exposed to child sexual abuse material (CSAM), focusing on the individual and organizational coping strategies, are presented in \cite{spence2023content}. In particular, this study highlights the importance of social support, role validation, and work-life separation, revealing a preference for mandatory, specialized therapy. \\
\textbf{Mitigation Strategies and Approaches}.
Employing image blurring to decrease the exposure of moderators to harmful data is studied in \cite{das2020fast}. However, blurring often compromises the conveyance of crucial image information, hampering a moderator's comprehension of the depicted content. Furthermore, the potential of grayscaling and blurring filters to minimize the emotional impact on content moderation workers is explored in \cite{karunakaran2019testing}. However, similar usability concerns, such as the obscurity of image content and eye strain are highlighted. Consequently, achieving a balance between preserving essential information and protecting viewers remains an ongoing challenge.

In addition to the professional contexts, image blurring has been employed by social media platforms such as Instagram\footnote{https://www.instagram.com/} to protect users from potentially disturbing content (content warning screens) \cite{bridgland2022meta,bridgland2022something}. However, several studies underscore the limitations of this method. A comprehensive analysis \cite{bridgland2023curiosity} related to Instagram's sensitive content screens found their efficacy in deterring users from accessing negative content to be low, even among individuals presenting with mental health issues. An effort of addressing the limitations of content warning screens suggests that providing additional information along with content warnings can reduce user engagement \cite{simister2023mind}. The underlying idea is that being informed about the content of an image can deter users from viewing the original image. Given these insights, the aim of this paper is to contribute to this field by exploring the utilization of AI filters for mitigating the effects of viewing disturbing imagery. By comparing these advanced AI approaches with conventional methods, we aim to deepen our understanding of this field and help devise more effective solutions to protect the mental well-being of individuals professionally required to interact with disturbing digital content.

\section{Methodology}
\label{sec:method}
A detailed description of the methodology, including the technical details, study format, and study distribution, is outlined in this Section.

\subsection{Image Style Transfer Algorithm}

Central to our methodology is the use of the Progressive Attentional Manifold Alignment (PAMA) \cite{luo2022progressive} style transfer algorithm, which operates on the premise of aligning the content manifold to the style manifold. This is a sophisticated, three-staged process that involves a channel alignment module, an attention module, and a spatial interpolation module. Each module serves a distinct yet interconnected function. The channel alignment module focuses on related content and style semantics, the attention module is responsible for establishing correspondence between features, and the spatial interpolation module then adaptively aligns the manifolds.
One of the key characteristics of PAMA is its capacity to alleviate the often-encountered style degradation problem, thus generating stylization outcomes that achieve state-of-the-art quality. In particular, PAMA offers regional consistency, content preservation, and high style quality. The inputs into the algorithm include the style image and the image set to be transformed.

Our study adopts three specific styles for transformation: grayscale drawing, slightly colored drawing, and painting. The grayscale drawing imparts a monochrome filter to the imagery, simplifying the visual content while preserving essential structural information. The slightly colored drawing adds a minimal amount of color, providing additional visual clues. The painting style transforms the image into a Renaissance rendition, further distancing the viewer from the graphic reality of the content. Finally, as regards the conventional filters, we employed two blurring filters. The first one applies the blur across the entirety of the image, whereas the second one selectively blurs only the portion of the image that contains the disturbing content.

\subsection{Study Design}

The first user-study segment was dedicated to profiling the participants. This preliminary part of the study comprised typical demographic questions and two profile-building questions. The latter aimed to indicate the participants' frequency of exposure to potentially disturbing UGC and their level of comfort or discomfort when exposed to graphic imagery. 

The second segment constitutes the core of the study. It was carefully structured to include two main phases, the first of which involved five transformed images—one for each filter under consideration (i.e., three AI-based and two traditional filters).
In this phase, participants were asked to rate a select subset of emotions from the Positive and Negative Affect Schedule (PANAS) scale. These emotions were carefully chosen for their relevance to the disturbing nature of the images - Distressed, Upset, Scared, Irritable, Nervous, Jittery, and Afraid. By rating these specific feelings after viewing each transformed image, participants were able to provide an empirical measure of their affective response, thus giving us an understanding of the emotional impact each filter had.  The same procedure was followed for the original images to establish the baseline emotional reactions. 
In addition to this emotion rating, participants were asked to engage in an interpretative exercise. They were prompted to provide a free text description of what they believed each image depicted. 
This exercise allowed us to determine how successfully each filter retained the necessary information. In the second phase of image filter evaluation, participants were shown four more image sets. Each set contained five variations of a single image, showcasing the effects of each filter. Instead of focusing on specific emotions as in the first phase, participants were asked to rate the overall level of disturbance triggered by each (filtered) image. This aspect of the study was aimed at understanding the overall effectiveness of each transformation in mitigating the negative impact of the original images.

The final segment of the study was designed to utilize the collective expertise and insights of the participants. A general feedback question was posed to participants, inviting them to share their thoughts on how AI technology, and specifically the AI style transformation approaches, can contribute to protecting users from the negative impact of being exposed to graphic imagery. The intent was to gain insights that would aid in further refining our approach, bridging gaps, and possibly revealing new research directions. 

The questionnaire used in this study is available as supplementary material. Note that it contains disturbing content.

\subsection{Distribution of the Study}

The study was distributed to a diverse array of professionals whose roles often necessitate engagement with potentially disturbing content. To this end, personalized emails were sent to carefully chosen, targeted individuals, including researchers, journalists, investigators, fact-checkers, documentalists, editors, political scientists, and producers. This initiative led to responses from more than 42 organizations, amounting to a total of 86 participants.
In addition to the focused outreach to specific professionals, the study was also made accessible through an open call for participation. This initiative garnered an additional 21 responses from various professions, including forensic analysts, operations managers, sociologists, technologists, post-production supervisors, and systems engineers. This combination of targeted and open-call distribution strategies aimed to diversify the sample population, ensuring a comprehensive evaluation of the proposed approach's efficacy across different contexts and levels of exposure to disturbing content.

\section{Results}
\label{sec:results}
\subsection{Demographics and profiling questions}

Beginning with the demographics of the participants, there was a diverse group in terms of age distribution, as evidenced by Table~\ref{tab:age}. The largest proportion of respondents, 44.86\%, fell into the 30-45 age bracket, reflecting a participant pool primarily composed of mid-career professionals. This was followed by the 45-60 age group, representing almost a third of the sample at 29.91\%. Younger participants, aged 18-30, constituted 22.43\% of the sample, while those aged over 60 were least represented at 2.80\%. 

Looking at gender diversity, as outlined in Table~\ref{tab:gender}, the distribution was predominantly binary. Male participants accounted for over half of the total at 54.21\%, while females constituted 41.12\%. Non-binary individuals represented a smaller proportion at 3.74\%, and a minimal percentage of 0.93\% opted not to disclose their gender.

Regarding the frequency of exposure to potentially disturbing UGC, as reported in Table~\ref{tab:freq}, it was found that the largest portion of participants, namely 34.58\%, encountered such content multiple times  a week. Those who reported daily exposure constituted 22.43\%, closely trailed by respondents who encounter disturbing material several times a month (21.50\%). A lesser proportion, 16.82\%, came across such content several times a year, while 4.67\% of the participants almost never encountered disturbing material online.
\begin{table}[t]
\centering
\caption{Age distribution.}
\begin{tabular}{cc}
\toprule
Age group &  Percentage \\
\midrule
18-30 & 22.43\% \\
30-45 & 44.86\% \\
45-60 & 29.91\% \\
$>$60 & 2.80\% \\
\bottomrule
\end{tabular}
\label{tab:age}
\end{table}
\begin{table}[t]
\centering
\caption{Gender distribution.}
\begin{tabular}{lc}
\toprule
Gender &  Percentage \\
\midrule
Male & 54.21\% \\
Female & 41.12\% \\
Non-binary & 3.74\% \\
Prefer not to say & 0.93\% \\
\bottomrule
\end{tabular}
\label{tab:gender}
\end{table}
\begin{table}[h]
\centering
\caption{Frequency of exposure to potentially disturbing UGC.}

\begin{tabular}{lc}
\toprule
Frequency & Percentage \\
\midrule
Almost never & 4.67\% \\
Several times a year & 16.82\% \\
Several times a month & 21.50\% \\
Several times a week & 34.58\% \\
Daily & 22.43\% \\
\bottomrule
\end{tabular} 
\label{tab:freq}
\end{table}
\begin{table}[t]
\centering
\caption{Self-perceived reactions to exposure of potentially graphic imagery.}
\begin{tabular}{lc}
\toprule
Response & Percentage \\
\midrule
Graphic imagery does not affect me negatively &  4.67\% \\
I rarely react negatively & 36.45\% \\
I sometimes react negatively & 39.25\% \\
I often react negatively & 14.95\% \\
I almost always react negatively & 1.87\% \\
Other responses & 2.79\% \\
\bottomrule
\end{tabular}
\label{tab:exp}
\end{table}

Table~\ref{tab:exp} presents the distribution of self-perceived reactions to exposure to graphic imagery. The 39.25\% of the participants indicated they sometimes react negatively to such content, while a slightly smaller proportion, 36.45\% reported rarely reacting negatively. Those who regularly had negative reactions comprised 14.95\% of the sample. A small fraction of 4.67\% indicated that graphic imagery does not affect them negatively. Only 1.87\% of participants claimed they almost always react negatively to such imagery, with a few respondents, i.e., 2.79\%, providing other responses.

\begin{table}[b]
\centering
\caption{Painting Style: Feelings while watching the image. The rating scale ranges from 1 (low) to 5 (high).}
\begin{tabular}{cccc}
\toprule
Feeling & Filtered & Original & Mitigation\\
\midrule
Distressed & 1.729 $\pm$ 0.907 & 3.122 $\pm$ 1.178 & 44.63\% \\
Upset & 1.439 $\pm$  0.826 & 2.833 $\pm$ 1.295 & 49.20\% \\
Scared & 1.458 $\pm$  0.872 & 2.061 $\pm$ 1.299  &  29.26\% \\
Irritable & 1.421 $\pm$ 0.847 & 1.949 $\pm$ 1.205  & 27.09\% \\
Nervous & 1.402 $\pm$ 0.775 & 2.122 $\pm$ 1.310 & 33.93\% \\
Jittery & 1.262 $\pm$ 0.649 & 2.163 $\pm$ 1.298 & 41.65\% \\
Afraid & 1.364 $\pm$ 0.719 & 1.990 $\pm$ 1.343 & 31.46\% \\
\midrule
mean & 1.439  $\pm$ 0.649 & 2.322 $\pm$ 1.065 &  38.03\% \\
\bottomrule
\end{tabular}
\label{tab:painting}
\end{table}
\begin{table}[t]
\centering
\caption{Colored Drawing Style: Feelings while watching the image. The rating scale ranges from 1 (low) to 5 (high).}
\begin{tabular}{cccc}
\toprule
Feeling & Filtered & Original & Mitigation \\
\midrule
Distressed & 1.374 $\pm$ 0.694 & 1.970 $\pm$ 1.096  & 30.25\% \\
Upset & 1.346 $\pm$ 0.754 & 1.680 $\pm$ 1.034  & 19.88\% \\
Scared & 1.308 $\pm$ 0.679 & 1.530 $\pm$ 0.948  & 14.51\% \\
Irritable & 1.252 $\pm$ 0.616 & 1.500 $\pm$ 0.948  & 16.53\% \\
Nervous & 1.327 $\pm$ 0.684 & 1.490 $\pm$ 0.959  & 10.93\% \\
Jittery & 1.243 $\pm$ 0.564 & 1.520 $\pm$ 0.948  & 18.22\% \\
Afraid & 1.355 $\pm$ 0.743 & 1.510 $\pm$ 0.987  & 10.26\% \\
\midrule
mean & 1.315  $\pm$ 0.570 & 1.603  $\pm$ 0.891  & 17.96\% \\
\bottomrule
\end{tabular}
\label{tab:colored_drawing}
\end{table}
\subsection{Trade-off between conveyed information and mitigation of negative feelings}
\label{subsec:part1}

As regards emotion alleviation, the Painting style filter illustrated a promising performance with an average negative feeling mitigation of 38.03\% as presented in Table~\ref{tab:painting}. The strongest mitigation effect was observed on feelings of being upset and distressed (i.e., the feelings that demonstrated the highest values w.r.t. the original image), registering a significant decrease of 49.20\% and 44.63\%, respectively. The least affected was the feeling triggered less when viewing the original image (i.e., irritability), with a mitigation rate of 27.09\%. Although this reduction spectrum suggests the potential of the Painting filter in diminishing the overall emotional distress incited by graphic images, it was not without its drawbacks. While 87 out of 107 participants (i.e., 81.31\%) were able to describe the content of the image, the provided responses revealed that the inherent abstraction of the Painting filter occasionally added an extra layer of distress, while some participants compared it to a piece of disturbing art. For instance, one of the responses was: `\textit{An injured person (though it is very unclear, and that's what makes it a bit disturbing)}'. This unintended consequence indicates that while the Painting style filter has a definite potential in reducing negative emotional reactions, it may unintentionally introduce certain elements of unease.

Furthermore, Table~\ref{tab:colored_drawing} presents the results for the Colored Drawing filter, indicating an average emotional mitigation of 17.96\%. The highest mitigation was observed for feelings of distress, 30.25\%, while feeling of fear saw the least mitigation, i.e., 10.26\%. It is worth noting that the original image was less disturbing (i.e., approximately 1.6 on the 1-5 rating scale) among the images involved in this study, which justifies the relatively low emotional mitigation (i.e., 17.96\%).
Although a total of 88 participants (i.e., 82.24\%) could comprehend the image, some participants reported difficulties in identifying specific objects or elements within the image.

Table~\ref{tab:drawing} shows that the Drawing style filter particularly excelled in preserving the interpretability of the image and mitigating the negative feelings. A majority of participants (i.e., 97.19\% or 104 out of 107) successfully identified several details, suggesting that this style maintained a high level of clarity. For example, one of the responses was the following: `\textit{A dead man lying on the floor in front of two other people, one in Flipflops (so no soldiers, but private people)}'. The average reduction in negative emotions was significant, averaging 30.38\%. It is worth noting that the feelings most profoundly triggered by the original images, such as being upset and distressed, experienced the highest reduction, with 38.62\% and 37.66\%, respectively. 

\begin{table}[t]
\centering
\caption{Drawing Style: Feelings while watching the image. The rating scale ranges from 1 (low) to 5 (high).}
\begin{tabular}{cccc}
\toprule
Feeling & Filtered & Original & Mitigation \\
\midrule
Distressed & 1.748 $\pm$ 0.912 & 2.804 $\pm$ 1.213  & 37.66\% \\
Upset & 1.626 $\pm$ 0.906 & 2.649 $\pm$ 1.267  & 38.62\% \\
Scared & 1.439 $\pm$ 0.815 & 1.897 $\pm$ 1.262  & 24.14\% \\
Irritable & 1.449 $\pm$ 0.849 & 1.876 $\pm$ 1.235  & 22.76\% \\
Nervous & 1.421 $\pm$ 0.790 & 1.990 $\pm$ 1.311 &  28.59\% \\
Jittery & 1.430 $\pm$ 0.766 & 2.031 $\pm$ 1.311  & 29.59\% \\
Afraid & 1.393 $\pm$ 0.844 & 1.844 $\pm$ 1.292  & 24.46\% \\
\midrule
mean & 1.501  $\pm$ 0.766 & 2.156  $\pm$ 1.132 & 30.38\% \\ 
\bottomrule
\end{tabular}
\label{tab:drawing}
\end{table}
As regards the Partially Blurring filter, a significant majority of 103 participants (i.e., 96.26\%) could interpret the image but primarily relied on the unblurred regions. 
In addition, Table~\ref{tab:p_blur} reports emotional mitigation results, the Partial Blurring style filter had a mean mitigation score of 25.54\%. Similarly to the previous filters, it was most effective on feelings of distress and being upset, with reductions of 31.96\% and 30.78\%, respectively. The least impacted emotion was fear, with a mitigation of only 16.10\%.

\begin{table}[b]
\centering
\caption{Partial Blurring Style: Feelings while watching the image. The rating scale ranges from 1 (low) to 5 (high).}
\begin{tabular}{cccc}
\toprule
Feeling & Filtered & Original & Mitigation  \\
\midrule
Distressed & 2.299 $\pm$ 1.143 & 3.379 $\pm$ 1.178  &  31.96\% \\
Upset & 2.215 $\pm$ 1.182 & 3.200 $\pm$ 1.260  & 30.78\% \\
Scared & 1.692 $\pm$ 1.032 & 2.096 $\pm$ 1.329  & 19.27\% \\
Irritable & 1.626 $\pm$ 0.995 & 2.232 $\pm$ 1.364  & 27.15\% \\
Nervous & 1.822 $\pm$ 1.156 & 2.295 $\pm$ 1.487  & 20.61\% \\
Jittery & 1.757 $\pm$ 1.071 & 2.404 $\pm$ 1.483  &  26.91\% \\
Afraid & 1.766 $\pm$ 1.194 & 2.105 $\pm$ 1.403  & 16.10\% \\
\midrule
mean & 1.883  $\pm$ 1.000 & 2.529  $\pm$ 1.169  & 25.54\% \\
\bottomrule
\end{tabular}
\label{tab:p_blur}
\end{table}

\begin{table}[t]
\centering
\caption{Blurring Style: Feelings while watching the image. The rating scale ranges from 1 (low) to 5 (high).}
\begin{tabular}{cccc}
\toprule
Feeling & Filtered & Original & Mitigation \\
\midrule
Distressed & 1.710 $\pm$ 0.858 & 2.773 $\pm$ 1.342  & 38.33\% \\
Upset & 1.607 $\pm$ 0.939 & 2.814 $\pm$ 1.294  & 42.89\% \\
Scared & 1.486 $\pm$ 0.817 & 1.701 $\pm$ 1.165  & 12.63\% \\
Irritable & 1.355 $\pm$ 0.717 & 1.835 $\pm$ 1.304  & 26.16\% \\
Nervous & 1.551 $\pm$ 0.838 & 1.794 $\pm$ 1.258  & 13.54\% \\
Jittery & 1.477 $\pm$ 0.872 & 1.794 $\pm$ 1.241  & 17.67\% \\
Afraid & 1.439 $\pm$ 0.815 & 1.670 $\pm$ 1.143 & 13.83\% \\
\midrule
mean & 1.518  $\pm$ 0.734 & 2.054  $\pm$ 1.111 & 26.09\% \\
\bottomrule
\end{tabular}
\label{tab:blur}
\end{table}
\begin{figure}[t]
\centering
  \includegraphics[width=0.6\linewidth, trim=1cm 0cm 2cm 1cm, clip]{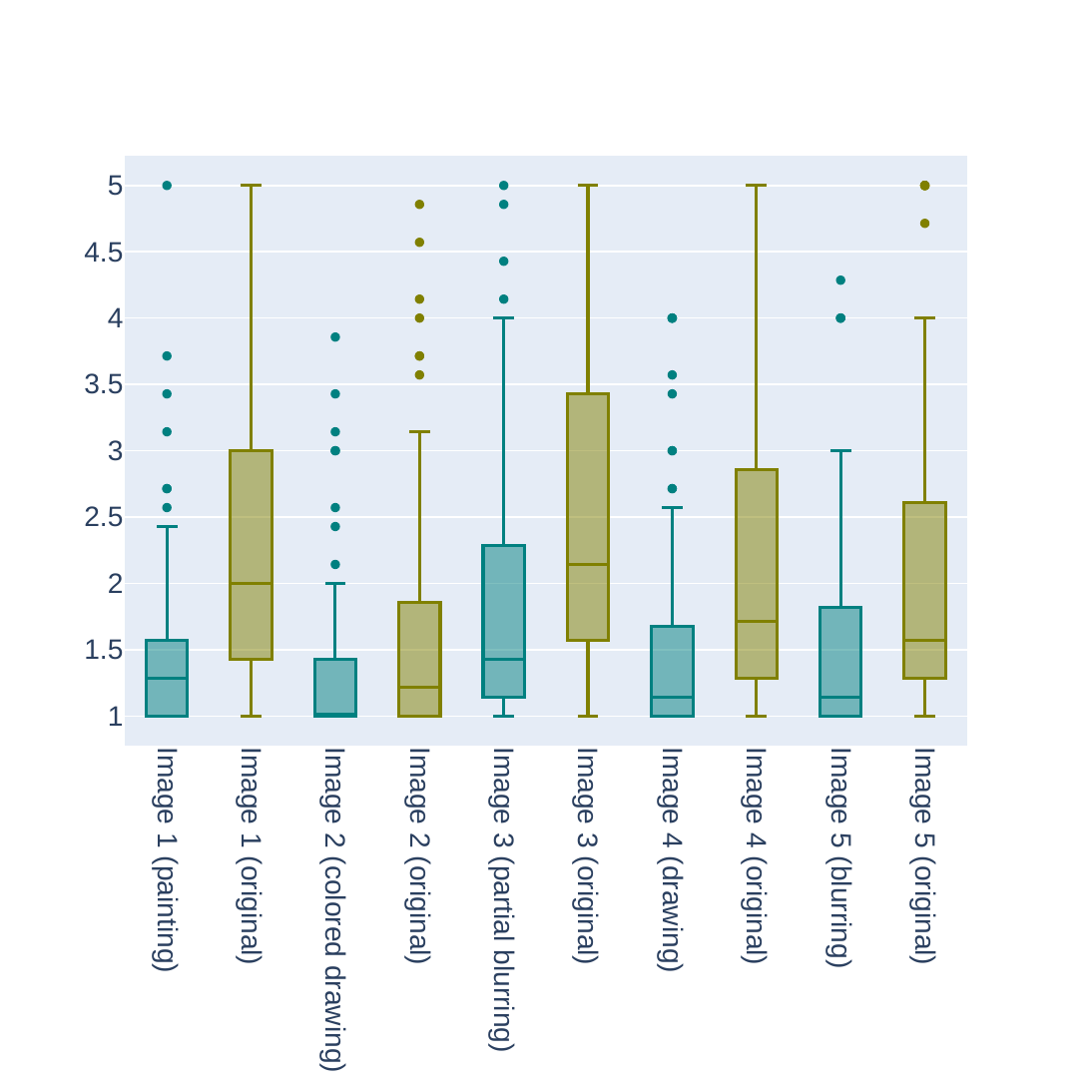}
\caption{Mean negative feelings for all filtered and original images.}
\label{fig:img_mean}
\end{figure}

\looseness=-1
In contrast to the other styles, the Blurring filter drastically affected image interpretability. Only a small fraction of participants (i.e., 6.54\%) could recognize the subject matter of the image, suggesting a high potential for information loss with this filter.
Regardless of its impact on interpretability, the Blurring style filter achieved a mean mitigation of 26.09\%. However, the significant information loss renders this filter less suitable for professionals where comprehending image details is crucial. Additionally, it is important to underscore that all discrepancies between the filtered and original images are statistically significant, affirmed by extremely small p-values ($<1\mathrm{e}{-10}$).
\looseness=-1

For further highlighting the discrepancies between filtered and original images, Figure~\ref{fig:img_mean} visualizes the mean negative feelings values reported in Tables~\ref{tab:painting}-\ref{tab:blur}. 
Overall, these findings endorse the Drawing style as the most effective filter in terms of maintaining a balance between interpretability and negative feelings mitigation. Its exceptional performance in preserving the image content while also significantly reducing negative feelings positions it as an optimal choice for professionals needing to interpret graphic images without undue emotional distress.


\subsection{Direct Filters Comparison}
To further explore the relative efficacy of different filters in mitigating the emotional impact of disturbing images, we involved four additional images in the study. Each image was subjected to all five filtering styles, and participants were asked to rate how disturbing they found the filtered images on a scale from 1 (not disturbing) to 5 (highly disturbing). The results of this investigation provide a direct comparison between the styles and allow us to examine the potential of AI filters in comparison to conventional approaches (i.e., blurring filters).

As presented in Table~\ref{tab:comp_styles}, based on mean disturbance ratings, the Drawing style filter was found to be the least disturbing with a mean score of 1.977 with a standard deviation equal to 0.733. The Colored Drawing and Painting style filters followed with scores of 2.439$\pm$0.799 and 2.692$\pm$0.804, respectively. The Partially Blurring and Blurring styles were perceived as the most disturbing, with mean scores of 3.371$\pm$0.926 and 3.002$\pm$0.873, respectively. These findings showcase high discrepancies among the filtering styles, with the AI-based Drawing style filter outperforming both the rest AI-based filters and the conventional blurring techniques.

Overall, the results above underscore the advantage of AI-based filters over traditional filters. However, as mentioned in Section~\ref{subsec:part1}, they should be considered in combination with the image interpretability, where the Drawing style offered the optimal trade-off across the evaluated filters.

\begin{table}[t]
\centering
\caption{Question: "How disturbing do you consider the following images?". Mean value across the 4 images. The rating scale ranges from 1 (not disturbing) to 5 (highly disturbing).}
\begin{tabular}{lcc}
\toprule
Style & Mean & Std \\
\midrule
Drawing Style & \textbf{1.977}  & 0.733 \\
Colored Drawing Style & 2.439  & 0.799 \\
Painting Style & 2.692  & 0.804 \\
Partially blurred & 3.371  & 0.926 \\
Blurred & 3.002  & 0.873 \\

\bottomrule
\end{tabular}
\label{tab:comp_styles}
\end{table}

\subsection{Practical Use and General Feedback}

To assess the practical usability of each filter, we asked the participants, `\textit{If the system you use in the scope of your work would provide the option to inspect images using this filter, to what extent would you use this option?}'. The responses, which ranged from 1 (would not use) to 5 (would use extensively), are compiled in Table~\ref{tab:use}.

With an average rating of 3.486, the Partial Blurring filter exhibits the greatest adoption rate among all filters. This is primarily attributed to its ability to blur only the distressing regions of an image, preserving crucial details. This aspect appears particularly beneficial to professionals who require comprehensive analysis of images in their work.
The full Blurring filter, on the other hand, garnered a lower mean score of 2.523, due to its tendency to conceal most of the image information.
Regarding the AI-based filters, the Drawing outperformed the Colored Drawing and Painting styles with a mean rating of 3.0 compared to 2.505 and 2.542, respectively. The preference for the Drawing filter can be attributed to its capacity to preserve the visual structure of the image while simultaneously distancing the viewer from the original scene. It is also worth noting that the black-and-white nature of the Drawing filter helps to distance viewers from the reality of the content, which is not the case for the Colored Drawing filter. 
From a technical standpoint, a standard framework incorporating such filters would encompass two AI models. The first one would be tasked with differentiating between potentially disturbing and safe content \cite{sarridis2022leveraging}, while the second model \cite{luo2022progressive} would then apply the proposed filters to the content classified as potentially disturbing by the first model.
\begin{table}[t]
\centering
\caption{Question: "If the system you use in the scope of your work would provide the option to inspect images using this filter, to what extent would you use this option?". The rating scale ranges from 1 (low) to 5 (high).}

\begin{tabular}{ccc}
\toprule
Style & Mean $\pm$ Std \\ 
\midrule
Blurring & 2.523 $\pm$ 1.231 \\
Partial Blurring & \textbf{3.486 $\pm$ 1.231} \\
\midrule
Painting & 2.542 $\pm$ 1.276 \\
Colored Drawing & 2.505 $\pm$ 1.239 \\
Drawing & \textbf{3.000 $\pm$ 1.259} \\

\bottomrule
\end{tabular} 
\label{tab:use}
\end{table}

Many participants highlighted that significant limitations
of such filters (both AI-based and conventional) exist – as it is often essential for professionals to view and investigate every minor detail in an original image, there were several suggestions on potential strategies to incorporate the proposed AI filters into their routine workflows.
The idea of using filters as a preparatory step before viewing the original image was brought up by several participants. By first viewing a filtered version, the viewer can prepare themselves emotionally for the impact of the real image, thus potentially reducing distress. This approach may be particularly effective in contexts where exposure to the original image is ultimately unavoidable, such as investigative journalism or forensics.
Furthermore, color was mentioned as a significant factor in perceiving images as disturbing. This aligns with previous psychological research suggesting that certain colors can evoke strong emotional responses \cite{kaya2004relationship}. Thus, adjusting the color palette of an image accordingly could be an effective way to reduce its emotional impact. 
Moreover, applying AI filters only to the regions of an image that depict disturbing content (as in the Partially Blurring filter) was another interesting suggestion. This targeted approach could maintain much of the image's original context and detail, while still protecting the viewer from the most distressing elements. 

In addition, the importance of variety and flexibility in filter options was emphasized by several participants. As user responses to different filters can vary widely based on individual sensitivities, having a range of filter styles to choose from could cover all individual requirements. 
Some participants also highlighted that AI filters could also prove particularly useful when dealing with large volumes of images.
Finally, the use of AI filters for repeated viewings of an image was noted. After the initial viewing of the original, filters can be applied in subsequent viewings to prevent the repeated experience of negative emotions. 

The following quotes are direct transcriptions from a subset of participants:
\begin{itemize}
    \item `\textit{Ultimately, in order to do an investigation, I will always eventually have to look at the original. With a technology as the one proposed, you advance a step from completely blurring (or overlaying) an image to giving the user some idea of what the image (original) may depict.}'
    \item `\textit{Those tests were really interesting and showed (to me) how much changing the color (especially the color red) makes an impact. So it would definitely help in my job (journalist/fact-checker) to have the possibility to use such filters by default. Sometimes we WILL have to look at the original picture, of course, if we need to investigate it further, but having a default filter making these less violent would be awesome. We would then only be forced to see the ones we need to investigate further.}'
    \item `\textit{I think the most important thing in limiting distress, for me personally, is that the photo allows me to have a symbolic understanding of what is happening without providing too many distinguishing characteristics.  The black-and-white line drawing method in particular seems excellent.  To that extent, I would be happy to use AI filters for researching gruesome topics if they allowed me to better understand information without suffering too many negative emotional effects.}'
    \item '\textit{I think it would be great to give these different options to journalists who are facing disturbing images and to let them choose the style they want to use (depending on the document they are looking at and depending on the way they react to those types of images - they need to be able to see original images if necessary of course). I think the best one is the drawing option in black and white, but maybe other styles would work better for other people. I would rather suggest only masking the zones which are graphic such as blood, wounds, and signs of starvation, instead of applying a new style to the whole image because it often suppresses any reality. Some filters on the whole image cartoonize it, making it look more as a contemporaneous artwork than some masked reality.}'
    \item '\textit{In some cases filters improve the content as they romanticize it in a special way. While in other cases they make the situation worse as they remove information and make you imagine whatever you want.}'
\end{itemize}

\section{Conclusion}
\label{sec:conc}
In this paper, we introduce a user study that investigates the potential of AI-based filters for mitigating the emotional impact caused by disturbing imagery, aiming to support professionals who regularly encounter such content in the context of their work or related activities.
The comprehensive study provided valuable insights into the effectiveness of different filter styles, with the Drawing style filter emerging as a particularly effective solution that maintains image interpretability while significantly reducing negative emotions.
Although limitations certainly exist, most notably the necessity for professionals to inspect every detail in the original images, the participants proposed potential strategies for integrating these AI filters into their workflows, such as utilizing AI filters as an initial, preparatory step to viewing the full image. Future studies can refine these filter techniques, test new ones, and experiment with the proposed integration methods to further optimize the balance between necessary exposure to critical content and the mitigation of its emotional impact. 
To conclude, there is a clear need for more research and activities in this domain. We hope that with our work we can contribute to reducing secondary or vicarious trauma of investigators, supporting the mental well-being of those who, because of the nature of their work and activities online, are exposed to graphic and potentially damaging imagery.

\section*{Ethics}
All participants were informed why the research is being conducted, whether or not anonymity is assured, and how the data they are collecting is being stored. We confirm that all the subjects have provided appropriate informed consent via the Google Forms platform. Finally, the ethics committee of the Centre for Research and Technology Hellas has granted ethical approval for this study.
\section*{Disclosure statement}
The authors report there are no competing interests to declare.

\section*{Acknowledgment}
This work was supported by the EU H2020 project MediaVerse under Grant Agreement 957252.

\bibliographystyle{unsrt}  
\small
\bibliography{references}  

\end{document}